\pdfoutput=1
\documentclass[conference]{IEEEtran}
\IEEEoverridecommandlockouts
\usepackage{amsmath,amssymb, amsfonts} 
\usepackage{url}
\usepackage{graphicx}
\usepackage[font={small}]{caption}
\usepackage{dblfloatfix}
\usepackage{algorithm}
\usepackage{algorithmic}
\usepackage{multirow}
\usepackage{color}
\def\BibTeX{{\rm B\kern-.05em{\sc i\kern-.025em b}\kern-.08em
    T\kern-.1667em\lower.7ex\hbox{E}\kern-.125emX}}

\begin{document}

\title{Learning and Generalizing Variable Impedance Manipulation Skills from Human Demonstrations*\\
\thanks{Yan Zhang, Fei Zhao and Zhiwei Liao are with The State Key Laboratory for Manufacturing System Engineering and Shaanxi Key Laboratory of Intelligent Robotics, School of Mechanical Engineering, Xi'an Jiaotong University, Xi'an Shaanxi, 710049, China.}

\thanks{This work was supported by the National Natural Science Foundation of China [Grant No. 52175029]
and the Department of Science and Technology of Shaanxi Province [Grant No. 2019ZDLGY14-07]}
\thanks{e-mail: ztzhao@xjtu.edu.cn}}

\author{Yan Zhang, Fei Zhao, Zhiwei Liao}
\maketitle

\begin{abstract}
By learning Variable Impedance Control policy, robot assistants can intelligently adapt their manipulation compliance to ensure both safe interaction and proper task completion when operating in human-robot interaction environments. In this paper, we propose a DMP-based framework that learns and generalizes variable impedance manipulation skills from human demonstrations. This framework improves robots$'$ adaptability to environment changes(i.e. the weight and shape changes of grasping object at the robot end-effector) and inherits the efficiency of demonstration-variance-based stiffness estimation methods. Besides, with our stiffness estimation method, we generate not only translational stiffness profiles, but also rotational
stiffness profiles that are ignored or incomplete in most learning Variable Impedance Control papers. Real-world experiments on a 7 DoF redundant robot manipulator have been conducted to validate the effectiveness of our framework.
\end{abstract}


\section{Introduction}\label{sec:introduction}
Intelligent robot assistants are increasingly expected to solve complex tasks in industrial environments and at home. Different from traditional industrial robots that complete a specific task in simple structured environments repeatedly, intelligent robot assistants may be required to perform different tasks in varying unstructured scenarios. Although Learning from Demonstration (LfD)
provides ordinary users with practical interfaces to teach robots manipulation skills\cite{billard2008survey}\cite{calinon2009robot}, an imitation learning framework with generalization ability is still needed to further reduce human intervention and to improve robots adaptation ability to environment changes. Besides, as robot assistants usually operate in human-populated environments, the manipulation compliance
should also be carefully scheduled to ensure safe interaction while robots completing the target task.

Dynamic Movement Primitives (DMP) model, firstly introduced in \cite{ijspeert2001trajectory} and further improved in
\cite{hoffmann2009biologically},\cite{ijspeert2013dynamical}, becomes popular in
the LfD community because of its powerful generalization ability. In general, DMP modulates each dimension of the
demonstrated movement trajectory as a second-order damped spring system. By approximating the non-linear force terms
and adjusting the attractor points, DMP can then generalize the demonstrated trajectory to similar situations while
keeping its overall shape. With this property, DMP has been used in various robot manipulation scenarios, such as
grasping objects at different positions\cite{hoffmann2009biologically}, playing drums at different heights\cite{hoffmann2009biologically}, and tennis swings \cite{ijspeert2002learning}. In these
papers, by elbowing robots with generalizable manipulation skills from once kinesthetic teaching, DMP greatly reduces human intervention during robot skill acquisition. However, the manipulation compliance is still ignored in most DMP-based skill learning framework.
\begin{figure*}[hbt]
    \centering
    \includegraphics[trim=0.2cm 0.2cm 0.2cm 0.3cm,clip, width=\textwidth, height=4cm]{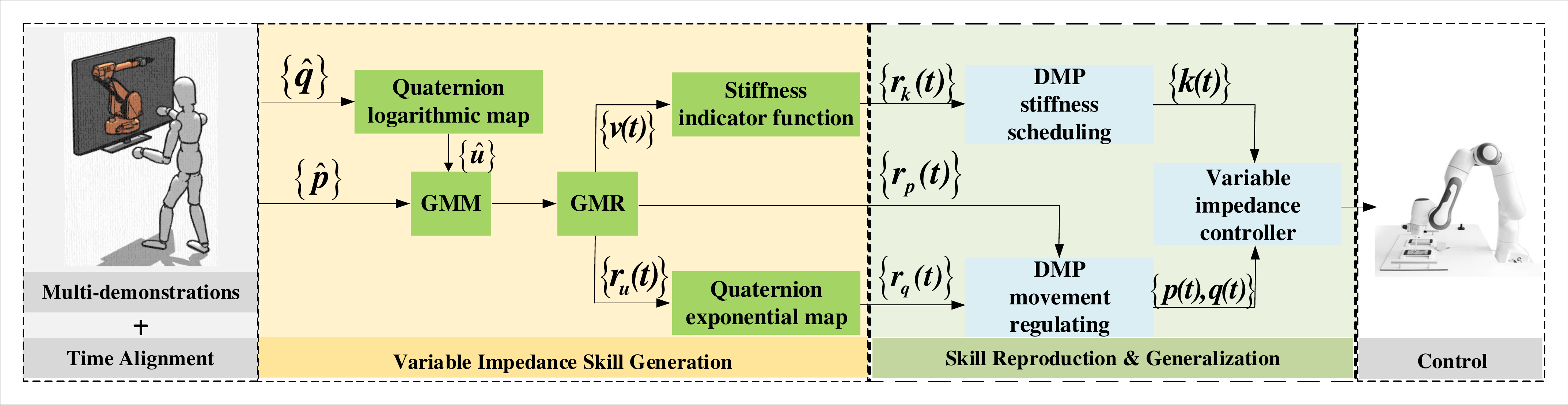}
    \caption{The overview of our proposed learning framework.}
    \label{fig:Overview of the framework}
    \vspace{-0.5cm}
\end{figure*}

Impedance Control (IC) \cite{hogan1985impedance} is commonly introduced in robot controller design for achieving complaint motions, in which
the controller can be viewed as a virtual spring-damp system between the robot end-effector and the environment. By
adapting the impedance parameters based on task requirements and environment dynamics, Variable Impedance Control (VIC)
can vary the manipulation compliance to ensure safe interaction and proper task completion \cite{ude2010task}. In
\cite{buchli2011learning}, the author
proposed a Reinforcement Learning (RL) framework titled Policy Improvement with Path Integrals (PI$^2$) where DMP was
firstly integrated with impedance parameters optimization. This RL framework parameterizes the movement trajectory and
impedance parameters with DMP models and then optimizes the parameters with the policy search optimization method.
Different from PI$^2$ where impedance parameters are learned indirectly, the authors in \cite{ajoudani2012tele} managed to explicitly estimate
human arm stiffness profiles based on the electromyographic (EMG) signals when human performing tasks. In \cite{wu2020framework}, we combined this EMG-based human arm stiffness estimation
method with optimal control theory and proposed an autonomous impedance regulation framework in a class of manipulation
tasks. While previously presented stiffness estimation methods show their potential in variable impedance manipulation skill acquisition, the complexity of estimating a set of parameters and the requirement of multiple EMG sensors make them inefficient and impractical for robot users.

In contrast, estimating impedance parameters from human demonstrations can be a more efficient way. In \cite{rozo2013learning}, the
authors proposed a human-robot collaborative assembly framework where stiffness matrix is estimated by Weighted
Least-Square (WLS) algorithms, with a small number of demonstrations and sensed force information. In \cite{abu2018force}, the authors
considered trajectory as a virtual spring-damper system, and estimated stiffness profiles based on demonstrated
kinesthetic trajectory and the associated sensed Cartesian force information. In \cite{calinon2010learning}, the authors also modeled
trajectory with a virtual spring-damper system, but they estimated the gain parameters of this spring-damper system
from demonstrated position trajectory to formulate a compliant controller. In their model, the demonstrated position
trajectory distribution is generated with the Gaussian Mixture Model-Gaussian Mixture Regression (GMM-GMR) algorithm
\cite{ghahramani1994supervised},\cite{calinon2007learning}, and the stiffness profiles are shaped so that the robot changes to a high stiffness in directions of low
variance. In \cite{kronander2012online}, this stiffness adaptation method is further applied to online decrease stiffness profiles when
humans perturbing the robot end-effector around its equilibrium point. However, most previously presented methods mainly focus
on estimating translational stiffness, rotational stiffness is ignored.

In this paper, we modified the stiffness estimation method in \cite{calinon2010learning} by integrating quaternion logarithmic
mapping function. This modification allows us to transform quaternions into decoupled 3D tangent vectors
and then to estimate rotational stiffness profiles based on the variances of tangent vectors. Besides, we noticed most works
in robot variable impedance manipulation skill learning focus on variable impedance skill reproducing and trajectory generalization, while the stiffness profiles are seldom generalized.
In this paper, we argue that generalizing stiffness profiles like trajectory can further improve robots' adaptability to
environment changes. To this end, we extend the classical DMP motion trajectory scheduling equations with stiffness generalization parts. The resulted learning framework is similar to the one presented in \cite{yang2018dmps} where joint stiffness profiles are estimated with EMG signals and then generalized in joint space. The main differences between us are that: 1) in our work, we estimate stiffness profiles from human demonstrations, which is more practical and efficient than EMG-based methods. 2) More importantly, we learn and generalize variable impedance manipulation skills in Cartesian space. This is a more natural way to regulate trajectory and stiffness profiles, as the goals for skill generalization
and the task constraints are normally presented in Cartesian space.

This paper is organized as follows. In Section 2, we introduce the methodology with the overview of our learning framework shown in Fig.\ref{fig:Overview of the framework}; In Section 3, we show the real-world validation experiments and analyze the results; In Section 4 and Section 5, we discuss and conclude this paper.

\section{Methodology}\label{sec:methodology}

As shown in Fig.\ref{fig:Overview of the framework}, our learning framework mainly consists of four parts: \textit{Trajectory
Collecting}, \textit{Variable Impedance Skill Generation}, \textit{Skill Reproduction and Generalization}, and\textit{
Real-world Robot Control}.

\textit{Trajectory Collecting}: a human demonstrator demonstrates to the robot how to accomplish one specific task
several times. Then, the demonstrated trajectories are collected and aligned into the same time scale.

\textit{Variable Impedance Skill Generation}: we transform the aligned quaternions  $\{\mathbf{\hat{q}}\}$ into tangent
vectors $\{\mathbf{\hat u}\}$ through the Quaternion Logarithmic Mapping Function. Then, we use GMM-GMR to encode both
$\{ \mathbf{{\hat u}}\}$ and $\{ \mathbf{{\hat p}} \}$ to obtain the trajectory distribution. The mean of
$\{ \mathbf{{\hat u}} \}$ is then transformed back to quaternions with the Quaternion Exponential Mapping Function. Meanwhile,
the variances of demonstrations are mapped to the reference stiffness profiles $\{ {\mathbf{r}_k}(t) \}$ with the Stiffness
Indicator Function.

\textit{Skill Reproduction and Generalization}: the extended DMP framework mainly consists of two parts: DMP movement
regulation block and DMP stiffness scheduling block. DMP movement regulation block generalizes the generated reference
pose trajectory $\{ {\mathbf{r}_p}(t), {\mathbf{r}_q}(t) \}$ to new scenarios and DMP stiffness scheduling block regulates
the reference stiffness profiles $\{ {\mathbf{r}_k}(t) \}$ to adapt to environment changes. Then, the torque commands
are calculated based on VIC equation with $\{{\mathbf{{r}_p}(t),\mathbf{{r}_q}(t)} \}$ and $\{ {\mathbf{{r}_k}(t)}\}$.

\textit{Real-world Control}: The calculated torque commands are then sent to real-world robots through Robot Operation System (ROS).

\subsection{Pre-processing}\label{subsec:preprocessing}
At first, N trajectories consist of positions and orientations of the end-effector are collected through kinesthetic
teaching. Each demonstration ${\mathbf{O}_i}(t){\rm{=}}\left\{{\mathbf{p}_i}({t_j}), {\mathbf{q}_i}({t_j}) \right\}$ ,
$ i=1,2,...,N$ is a ${M_i} \times7$ matrix, where ${M_i}$ indicates the total number of datapoints of the ${i^{th}}$
demonstrated trajectory, ${\mathbf{p}_i}({t_j}) = \left\{ {{p_{i,x}}({t_j}),{p_{i,y}}({t_j}),{p_{i,z}}({t_j})} \right\}$
and ${\mathbf{q}_i}({t_j}) = \left\{ {{q_{i,w}}({t_j}),{q_{i,x}}({t_j}),{q_{i,y}}({t_j}),{q_{i,z}}({t_j})} \right\}$
represent the position and unit quaternion of ${i^{th}}$ trajectory at ${t_j}$ timestep, respectively. Next, we
align the collected trajectories into the same time scale $\left[ {0,T} \right]$, for a given $T > 0$. This time
alignment process is done as follows: assume ${t_0}$ and ${t_1}$ be the initial and final time of a given trajectory
${\mathbf{O}_i}(t)$ . The aligned trajectory is then represented as:
\begin{equation}
{\mathbf{\hat O}_i}(t){\rm{ = }}{\mathbf{O}_i}(\frac{{T(t - {t_0})}}{{{t_1} - {t_0}}}),i = 1,2,...,N
\end{equation}
with ${\mathbf{{\hat O}}_i}(t){\rm{ = }}\left\{ {{{\mathbf{\hat p}}_i({t_j})},{{\mathbf{\hat q}}_i({t_j})}} \right\}$ .

\subsection{Variable Impedance Skill Generation}\label{subsec:Skill Generation}
\subsubsection{Quaternion Logarithmic and Exponential Mapping Functions}\label{subsubsec:quaternion maps}
Unlike positional part, there is no minimal and singularity-free representation for orientational part. The stiffness estimation
method in \cite{calinon2010learning} is effective to estimate translational stiffness profiles from position trajectories. However, it might not be
feasible to encode orientation trajectories and estimate rotational stiffness information directly. Inspired by
Quaternion-based DMP framework in \cite{ude2014orientation}\cite{pastor2011online}, where Quaternion Logarithmic Mapping Function is applied to calculate the 3D
decoupled distance vector between two quaternions, in this paper, we also utilize this mapping function to transform
quaternions into decoupled 3D vectors. The generated 3D vectors can be considered as the distance vectors between the transformed
quaternion and the original quaternion $\mathbf{1}{\rm{ = }}(1,0,0,0) $. Besides, Quaternion Exponential Mapping Function
is also presented here to transform the distance vectors back into quaternions for orientation trajectory encoding and
generalization with the extended DMP framework later.

Given an unit quaternion, the Quaternion Logarithmic Map Function $({S^3}\to{R^3})$ is written as:
\begin{equation}
    \mathbf{\hat u} = \log (\mathbf{{\hat q}}) = \left\{ \begin{array}{l}
        \arccos ({{\hat q}_w})\frac{{({{\hat q}_x},{{\hat q}_y},{{\hat q}_z})}}{{||{{\hat q}_x},{{\hat q}_y},{{\hat q}_z}||}},({{\hat q}_x},{{\hat q}_y},{{\hat q}_z}) \ne\mathbf{ \mathord{\buildrel{\lower3pt\hbox{$\scriptscriptstyle\rightharpoonup$}}
        \over 0}} \\
        (0,0,0),otherwise
    \end{array} \right.
    \label{con:quat_log}
\end{equation}

Correspondingly, the Quaternion Exponential Map Function $({R^3} \to {S^3})$ is defined by:
\begin{equation}
    \mathbf{\hat q} = \exp (\mathbf{\hat u}) = \left\{ \begin{array}{l}
        (\cos |\mathbf{\hat u}||,\frac{{\sin ||\mathbf{\hat u}||}}{{||\mathbf{\hat u}||}} \bullet \mathbf{\hat u}),\mathbf{\hat u} \ne (0,0,0)\\
        \mathbf{1}{\rm{ = }}(1,0,0,0),otherwise
    \end{array} \right.
    \label{con:quat_exp}
\end{equation}
where $\mathbf{\hat u }= ({\hat u_x},{\hat u_y},{\hat u_z}) \in {T_1}{S^3} \equiv {R^3}$ represents a tangent vector in
the tangent space ${T_1}{S^3}$. 

In Eq. \eqref{con:quat_exp}, the
exponential map transforms a tangent vector $\mathbf{\hat u}$ into a unit quaternion $\mathbf{\hat q}$, a point
in ${S^3}$ at distance $||\mathbf{\hat u}||$ from $\mathbf{1}$ along the geodesic curve beginning from $\mathbf{1}$ in
the direction of $\mathbf{\mathbf{\hat u}}$. Additionally, when we limit ${\rm{||}}\mathbf{\hat u}|{\rm{| < }}\pi {\rm{ }}$
and $\mathbf{\hat q }\ne  (- 1,0,0,0)$, these two mapping is continuously differentiable and inverse to each other.

\subsubsection{Stiffness Indicator Function}
\label{subsubsec:Stiffness Indicator Function}
The stiffness indicator function maps the variances of demonstrated trajectories to stiffness profiles. The basic idea is that: in the region where demonstrated-trajectory has a low variance, the robot should keep at a high stiffness level to track the reference trajectory precisely. While for the high variance parts, the robot can keep at a relatively low stiffness level to ensure manipulation compliance.
To generate relatively lower stiffness profiles, we apply the left part of a quadratic function with a positive quadratic coefficient as the stiffness indicator function:
\begin{equation}
    {k_l}(t) = {a_l}{({d_l}(t) - {d_{l}^{max}})^2} + {k_{l}^{min}}
    \label{con:indicator}
\end{equation}
\[{a_l} = \frac{{k_l^{\max } - k_l^{\min }}}{{{{(d_l^{\min } - d_l^{\max })}^2}}} > 0\]
where $k_l^{\min },k_l^{\max }$ are the minimal and maximal translational or rotational stiffness values in direction
$l \in \left\{ {x,y,z} \right\}$, given based on the robot hardware limitations and real-world task constraints. Besides, $d_l^{\min },d_l^{\max }$ indicate the minimum and maximum of the standard deviation of the demonstrated trajectories in direction $l$.

\subsection{Skill Reproduction and Generalization}\label{subsec:skill_generalization}
\subsubsection{Extended DMP Model}\label{subsubsec:dmp}
DMP model considers a trajectory as a second-order damped spring system with a non-linear force term $f(\bullet)$,
like Eq.\eqref{con:dmp_pos}. Given a demonstrated trajectory, by solving the regression problem of the non-linear force term, DMP
can imitate this trajectory and generalize it to new similar scenarios by adjusting the goals. However, when transferring human skills to robots, most classical DMP only encodes pose trajectories, which may lose part of the compliance of demonstrated skills. To learn and generalize compliant variable impedance manipulation skills, we extend the original DMP model by integrating the stiffness scheduling equations
in Eq.\eqref{con:dmp_stiffness}. Meanwhile, Quaternion-based DMP, Eq. \eqref{con:dmp_quat} is united in our extended model to
encode the reference orientation trajectory.
\begin{equation}
    \tau \mathbf{y} = \alpha_{p}(\beta_{p}(\mathbf{p}_{g}-\mathbf{p})-
    \mathbf{y})+\mathbf{f}_{p}(x)
    \label{con:dmp_pos}
\end{equation}
\begin{equation}
    \tau\mathbf{z} = \alpha_{k}(\beta_{k}(\mathbf{k}_{g}-\mathbf{k})-\mathbf{z})+\mathbf{f}_{k}(x)
    \label{con:dmp_stiffness}
\end{equation}
\begin{equation}
    \tau \boldsymbol{\dot{\eta }}={{\alpha }_{q}}({{\beta }_{q}}2\log ({\mathbf{{q}_{g}}}*\mathbf{\bar{q}})-
    \boldsymbol{\eta} )+{\mathbf{{f}}_{q}(x)}
    \label{con:dmp_quat}
\end{equation}
\begin{equation}
    \tau \mathbf{p}=\mathbf{y}
\end{equation}
\begin{equation}
    \tau \mathbf{k}= \mathbf{z}
\end{equation}
\begin{equation}
    \tau \mathbf{\dot{q}}=\frac{1}{2}\boldsymbol{\eta} *\mathbf{q}
    \label{con:quat_deriv}
\end{equation}
where $\mathbf{p},\mathbf{p}_g \in {R^3}$ indicate current position of the robot's end-effector in Cartesian space and the final goal position; $\mathbf{k},\mathbf{k}_g \in {R^6}$
represent the main diagonal elements of stiffness matrix and their target values, respectively;
$\mathbf{q},\mathbf{q}_g \in {S^3}$ are robot's current orientation and the final goal orientation; $ {\alpha _p},{\alpha _k},
{\alpha _q},{\beta _p},{\beta _k},{\beta _q}$ are constant parameters; $\tau $ indicates the time scaling factor
that is used to adjust the duration of the task; $\mathbf{y},\mathbf{z}$ represent position
velocity, the derivative of stiffness, and the tangent vector calculated by the quaternion logarithmic map in
Eq. (2); $\boldsymbol{\dot q}$ is the quaternion derivative that satisfies Eq.\eqref{con:quat_deriv}, where $\boldsymbol{\eta}$ is the angular velocity; Besides,
$\mathbf{\bar q}$ denotes the quaternion conjugation, with the definition: $\mathbf{\bar q} = ({q_w}, - {q_x},
- {q_y}, - {q_z}) $. Finally, the symbol $ * $ indicates the quaternion product.

The whole extended DMPs model is synchronized by the canonical system:
\begin{equation}
\tau \dot x =  - {\alpha _x}x
\end{equation}
where $x$ is the phase variable to avoid explicit time dependency; ${\alpha _x}$ is a positive constant and $x(0)$ is set as 1.

The non-linear forcing terms ${\mathbf{f}_p}(x),{\mathbf{f}_q}(x),{\mathbf{f}_k}(x)$ are functions of $x$ and can be
regressed with Locally Weighted Regression (LWR) algorithm:
\begin{equation}
\mathbf{f}(x) = \frac{{\sum\nolimits_{s = 1}^S {{\boldsymbol{\theta} _s}{\psi _s}({x_j})} }}{{\sum\nolimits_{s = 1}^S
{{\psi _s}({x_j})} }}x
\end{equation}
where $\mathbf{f}(x)$ represents ${\mathbf{f}_p}(x),{\mathbf{f}_q}(x),{\mathbf{f}_k}(x)$ in general. $S$ is the number
of radial basis functions used. 

Given demonstrated trajectories, S-column parameter matrix $\boldsymbol{\theta} $ can be obtained by solving the following equations:
\begin{equation}
\mathbf{f}_{p}({x_j})= \mathbf{G}_{p}^{ - 1}(\tau^2{\mathbf{p}_j}+\tau\alpha_{p}({\mathbf{p}_j}-\alpha_{p}\beta_{p}({\mathbf{p}_{g}}-{\mathbf{p}_j}))
\end{equation}
\begin{equation}
\mathbf{f}_{k}({x_j})= \mathbf{G}_{k}^{ - 1}(\tau^2{\mathbf{k}_j}+\tau\alpha_{k}({\mathbf{k}_j}-\alpha_{k}\beta_{k}({\mathbf{k}_{g}}-{\mathbf{k}_j}))
\end{equation}
\begin{equation}
{\mathbf{f}_q}({x_j}) = \mathbf{G}_q^{ - 1}(\tau {\dot {\boldsymbol{\eta} _j} - {\alpha _q}({\beta _q}2\log
({\mathbf{q}_g} * \mathbf{\bar q}_j}) - {\boldsymbol{\eta}_j})
\end{equation}
\begin{equation}
{\psi _s}(x) = \exp ( - {h_s}{(x - {c_s})^2})
\end{equation}
where ${\boldsymbol{G}_{p}}=diag({\mathbf{p}_g}-{ {\mathbf{p}_{_{\rm{0}}}}})\in {R^{3 \times 3}}$,  ${\boldsymbol{G}_{k}} = diag({\mathbf{k}_g} -{{\mathbf{k}_{_{\rm{0}}}}})\in {R^{6 \times 6}}$, ${\mathbf{G}_q}=diag(2\log\mathbf{q}_g-\mathbf{q}_0)\in {R^{3\times 3}}$are spatial scaling factors.${h_s},{c_s}$ are the width
and center of Gaussian distribution ${\psi_s}(x)$.

\subsubsection{Variable Impedance Control}\label{subsubsec:vic}
With the scheduled pose trajectory $\left\{ {\boldsymbol{p},\boldsymbol{q}} \right\}$
and stiffness profiles $\boldsymbol{k}$ , we can calculate the command torques based on the variable impedance control
equation:
\begin{equation}
\mathbf{\Gamma}  = {\mathbf{J}^T}(\mathbf{K}\boldsymbol{e} + \mathbf{D}\boldsymbol{\dot{e}} ) + {\mathbf{\Gamma_{ff}}}
\end{equation}
where the stiffness matrix $\mathbf{K} = diag(\boldsymbol{k}) \in {R^{6 \times 6}}$ and damping matrix $\boldsymbol{D}=\sqrt{2\boldsymbol{K}}$; $\mathbf{J}$ is the Jacobian matrix, ${\mathbf{\Gamma}}$ represents the joint torque and $\mathbf{\Gamma}_{ff} $ is the joint torque contributed by feed-forward term.
${\mathbf{e}},{\mathbf{\dot{e}}}$ denote the reference pose trajectory tracking error and tracking velocity.

\section{Experimental Evaluation}
In this section, the 7DoF Franka Emika Robot(Panda) is used in our experimental study. For all the experiments, Panda
is controlled under libfranka scheme with 1kHz control frequency. A toy task, serving drinks is illustrated to show
that our framework 1) learns a reasonable variable impedance manipulation skill from human demonstrations; 2) enables the
robot to generalize the reference stiffness profiles to adapt to the changes of the robot end-effector.

As discussed in \cite{pastor2011online}, pouring the content of a bottle into a cup can be done with kinematic control model. However, in human-populated environments, a person may incautiously push the robot while it reaching for the cup. A stiff controller will respond with high forces, which may hurt the user and cause the liquid to spill. It would thus be desirable to control the way that the robot responds to translational and rotational perturbations.

In the real world, human tends to gradually increase the translational and rotational stiffness while reaching the cup and keep at high stiffness while pouring drinks. Besides, humans can also easily adapt to the shape and weight changes of the grasping bottle. In this experiment, we show that our framework learns reasonable variable impedance manipulation skills and also enables Panda to adapt to the weight and shape changes of the bottle like humans, by endowing it with variable impedance manipulation skill generalization ability.

\begin{figure}[htb!]
\centering
 \includegraphics[trim=0.2cm 0.2cm 0.2cm 0.3cm,clip,width=1\linewidth]{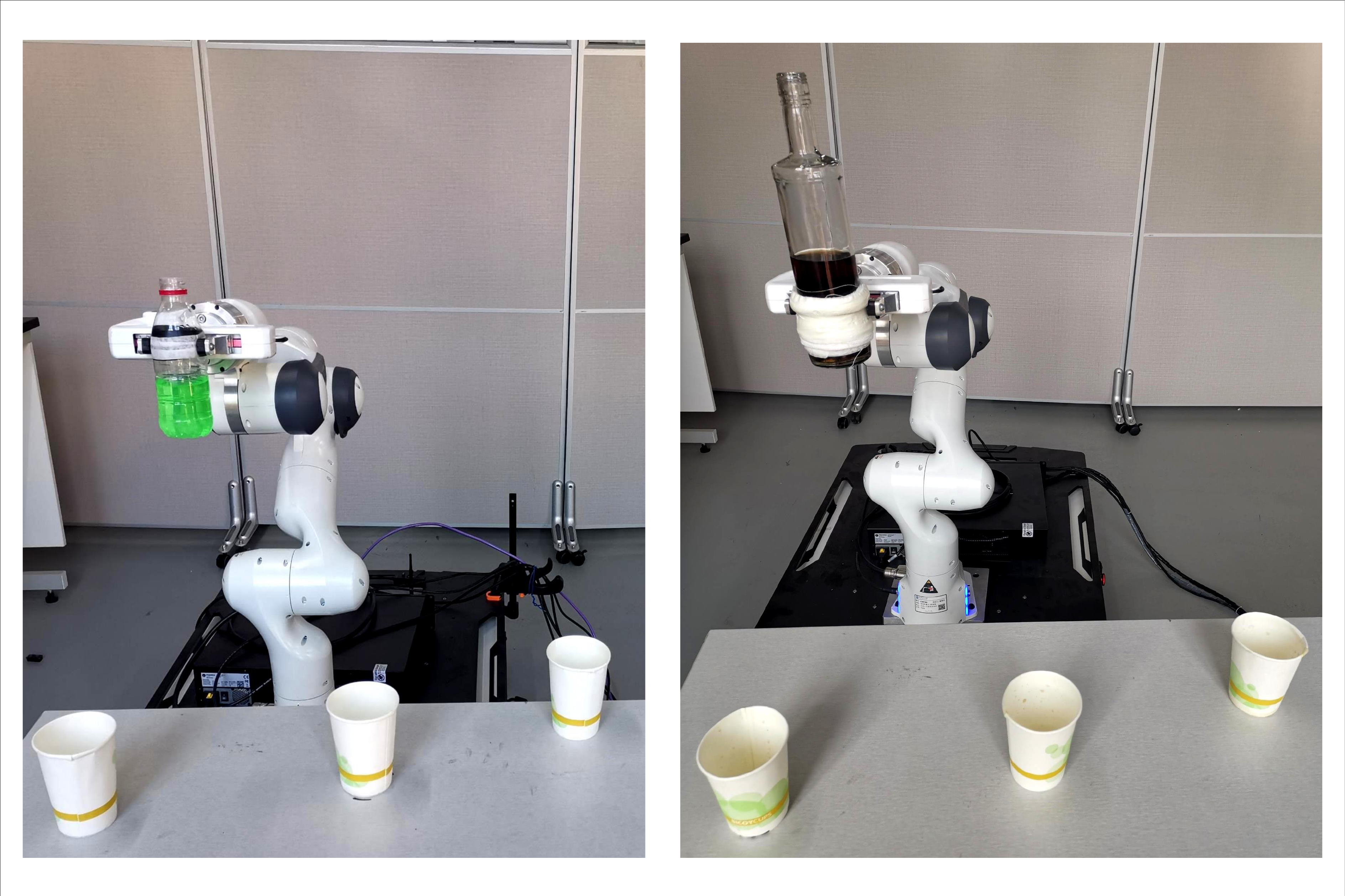}
    \caption{Experiment setup}
    \label{fig:exp_setup}
    \vspace{-0.5cm}
\end{figure}

The experiment set up is shown in Fig.\ref{fig:exp_setup}. Panda is expected to 1) pour water from a 0.25 kg plastic bottle into the middle cup on the table by reproducing the demonstrated pose trajectory and stiffness profiles; 2) pour water into the other two cups by generalizing the reference pose trajectory; 3) pour wine from a 0.9 kg glass bottle into the cups by generalizing the reference trajectory and stiffness profiles simultaneously.
\begin{figure}[htb!]
  \centering
    \includegraphics[trim=0.2cm 0.2cm 0.2cm 0.3cm,clip,width=1\linewidth]{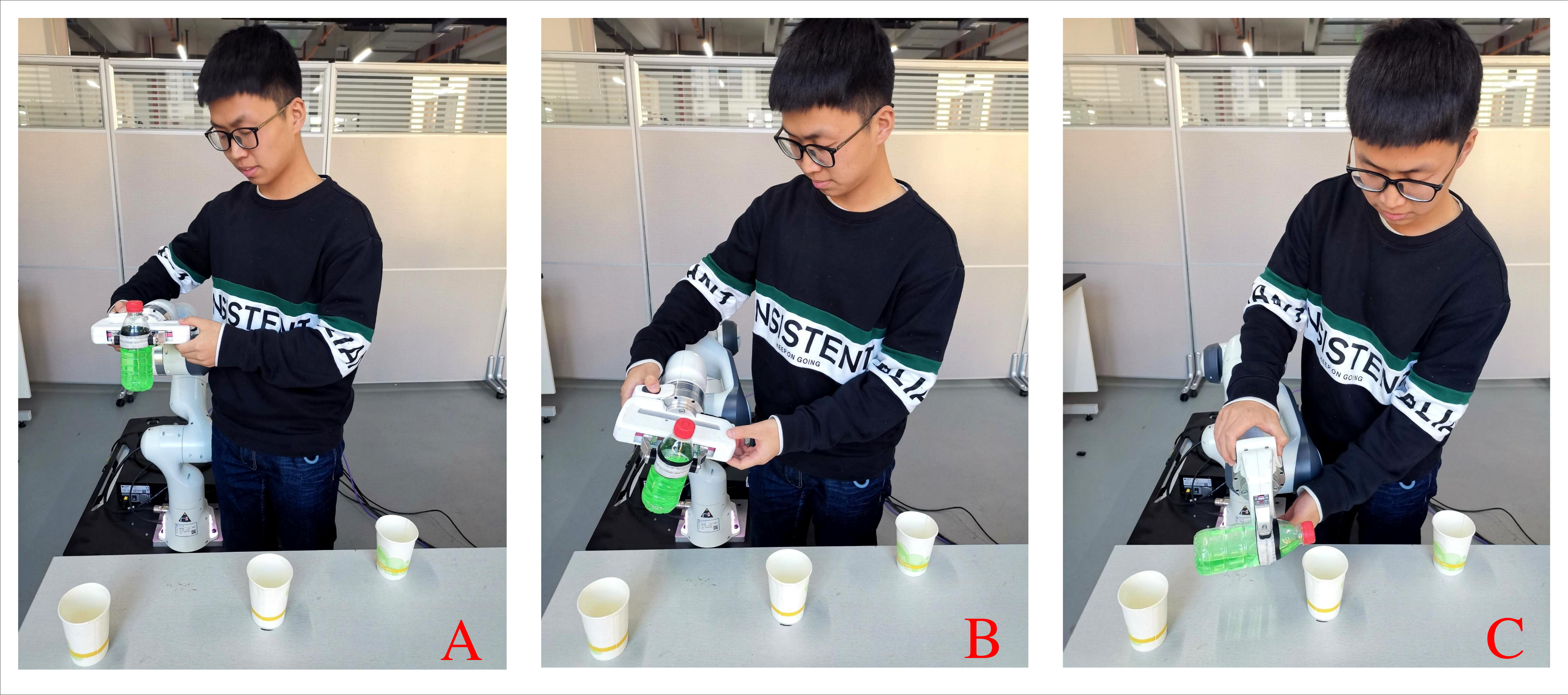}
    \caption{Kinesthetic teaching.The robot user demonstrates Panda how to pour water into the middle cup.}
    \label{fig:teaching}
    \vspace{-0.3cm}
\end{figure}

\subsection{Learning Variable Impedance Manipulation Skill}\label{subsec:skill learning}

\begin{figure}[hbt!]
    \centering
    \includegraphics[trim=0.3cm 0.3cm 0.5cm 0.5cm,clip,width=1\linewidth]{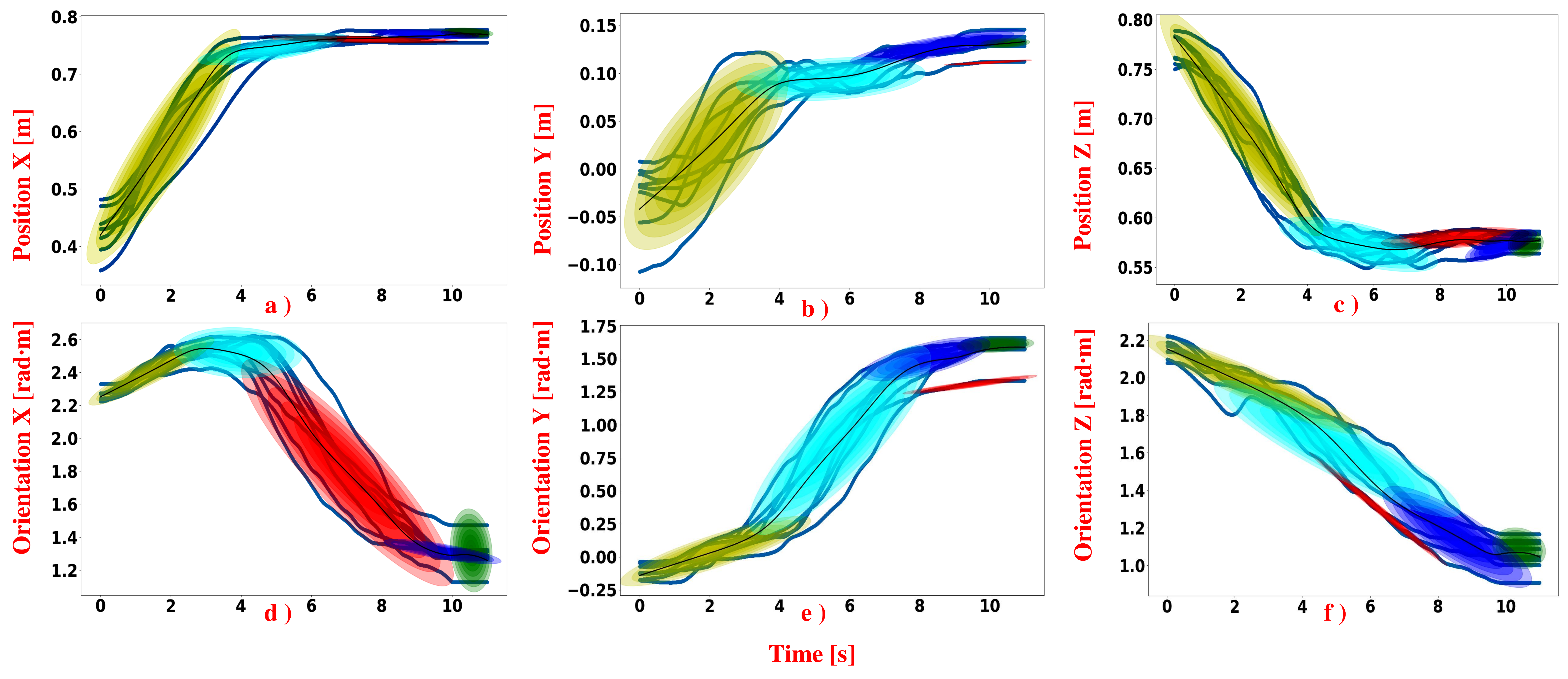}
    \caption{GMM-GMR encodes the positional and orientational datapoints.The demonstrated trajectories, the estimated mean functions, and the trained Gaussian kernels are marked with blue lines, black lines, and colorful ellipses, respectively.}
    \label{fig:trajectory processing}
    \vspace{-0.3cm}
\end{figure}

We first show Panda how to pour water into the middle cup on the table 8 times in slightly different situations, with
kinesthetic teaching (Fig.\ref{fig:teaching}). Then, the collected pose trajectories were aligned into the same time scale T = 11 seconds.
Next, we transformed the unit quaternions into tangent vectors with the Quaternion Logarithmic Mapping Function, Eq.\eqref{con:quat_log}, and encoded both positional and orientational datapoints with GMM-GMR, with H=6 Gaussian components, shown in Fig.\ref{fig:trajectory processing} Finally, the mean tangent vectors in Fig.\ref{fig:trajectory processing} (d-f) were converted back into unit quaternions through
the Quaternion Exponential Mapping Function, Eq.\eqref{con:quat_exp}.

\begin{figure}[hbt!]
    \centering
    \includegraphics[trim=0.3cm 1.0cm 0.3cm 0.3cm,clip,width=0.9\linewidth]{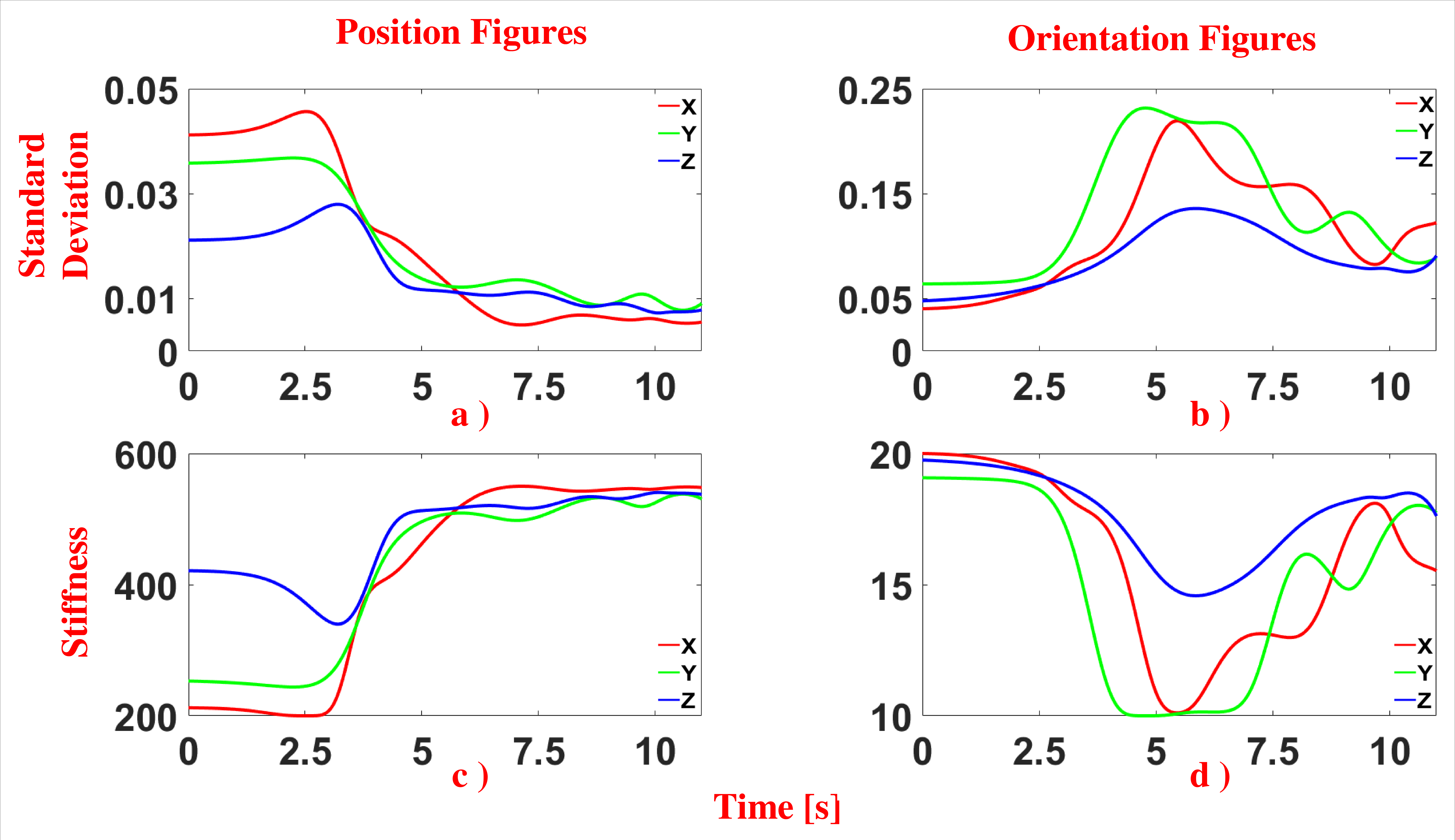}
    \caption{a-b) represent the standard deviation of the collected motion trajectories; c-d) show the estimated stiffness profiles}
    \label{fig:stiffness mapping results}
    \vspace{-0.3cm}
\end{figure}

The reference stiffness profiles is estimated with our Stiffness Indicator Function, Eq.\eqref{con:indicator}, based on the variances of the demonstrated trajectories. For pouring water experiment, the minimal and maximal translational stiffness values allowed are set as $k^{\min }= 200N/m$ and $k^{\max } = 550N/m$, and the values for the rotational stiffness are $k^{\min } = 10N/(rad \bullet m)$, $k^{\max}= 20N/(rad \bullet m)$. The estimated stiffness profiles are presented together with the standard deviations of the collected trajecotries in Fig.\ref{fig:stiffness mapping results}.

Overall, the estimated translational stiffness profiles increase when the robot reaching for the cup and keep high when pouring water. Before beginning the experiment, we expect the rotational stiffness profiles show the same tendency to the translational part. Nevertheless, they actually start from high values, decrease gradually at around 3s and keep increasing to high values from around 7s until the end. This tendency is actually more reasonable than what we expected. At the beginning phase of pouring water, we unconsciously perform high rotational stiffness to avoid our hand rotate to prevent the water from spilling out. Therefore, our framework can indeed generate reasonable stiffness profiles from human demonstrations.

\begin{figure}[hbt!]
    \centering
    \includegraphics[trim=0.5cm 1.5cm 0.5cm 0.3cm,clip,width=1\linewidth]{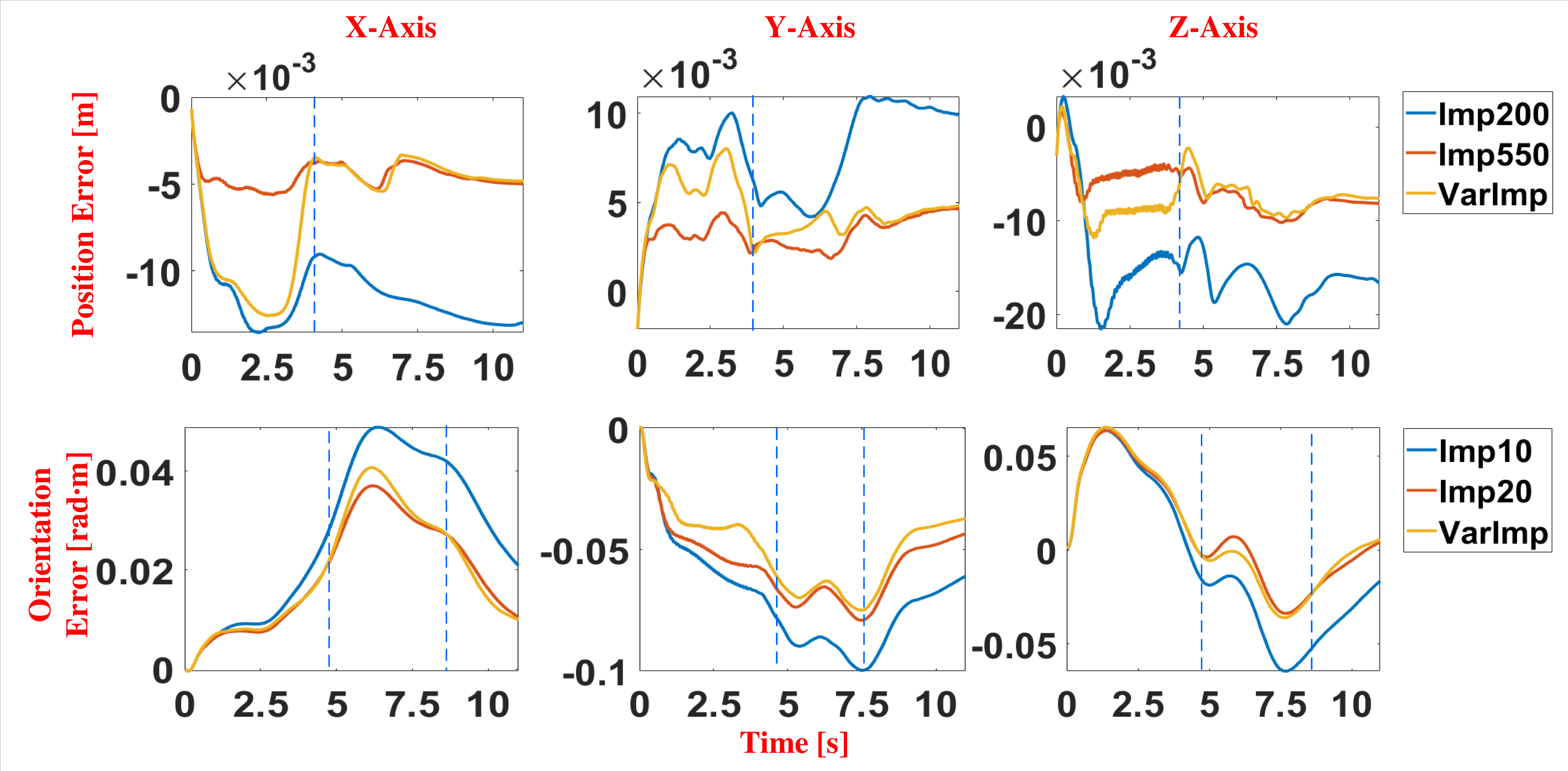}
    \caption{Comparison of mean tracking pose errors in different stiffness modes}
    \label{fig:real-world experiment 2}
    \vspace{-0.3cm}
\end{figure}

To illustrate we successfully transferred the stiffness features to Panda in the real world, we recorded the mean tracking errors of pouring water into the middle cup with 1)the estimated stiffness profiles, 2)the minimal stiffness value allowed, 3) the maximal stiffness value allowed. The result is shown in Fig.\ref{fig:real-world experiment 2}

We firstly set rotational stiffness at $20 N/(rad\bullet m)$, then used $200 N/m$, $550 N/m$ and the reference translational stiffness profiles to accomplish the pouring water experiment for 3 times for each translational stiffness mode. The mean positional
errors were shown in the upper 3 graphs in Fig.\ref{fig:real-world experiment 2}. It is noticeable that our variable
impedance controller behaves like a 200 N/m constant stiffness controller from 0s to around 4s, as the yellow lines are
close to the blue lines during this period. After 4s, the yellow lines almost coincide with the red lines which represent the tracking errors of the 550 N/m constant stiffness controller. This exactly reflects the tendency of the translational stiffness in Fig.\ref{fig:stiffness mapping results}$c)$. As for the rotational part, we set the translational stiffness at 550 N/m, and tested the mean tracking errors of $10N/(rad\bullet m)$, $20N/(rad\bullet m)$, and the reference orientational stiffness profiles. The results are shown in the lower 3 graphs in Fig.\ref{fig:real-world experiment 2}. The results also inflect the overall tendency of the orientational stiffness in Fig.\ref{fig:stiffness mapping results}$d)$. Therefore, with our extended-DMP model, Panda learns reasonable variable impedance manipulation pouring water skill.

\subsection{Variable Impedance Skill Generalization}\label{subsec:skill generalization}
\begin{figure}[hbt!]
    \centering
    \includegraphics[trim=0.3cm 1.0cm 0.1cm 0.3cm,clip,width=0.99\linewidth]{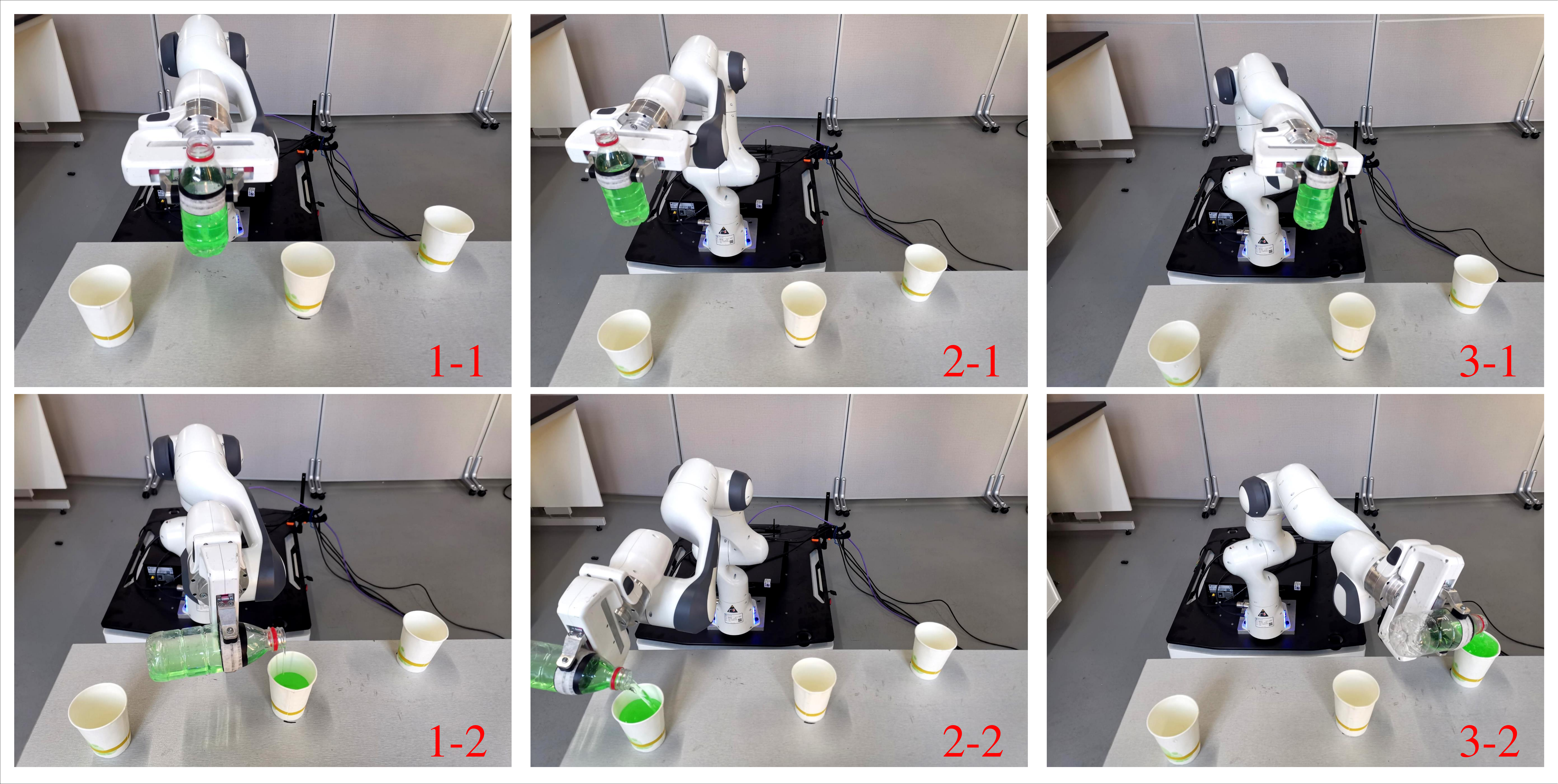}
    \caption{Shortcuts of real-world pouring water experiment}
    \label{fig:real-world experiment 1}
    \vspace{0.3cm}
\end{figure}

In this section, we generalize the learned variable impedance manipulation skill to new scenarios and show our extended-DMP model further improves Panda$'$s adaptability to the shape and weight changes of the grasping bottle. First, we used the reference stiffness profiles and generalized the pose trajectory to show our framework inherits the motion generalization ability of DMP model. As shown in Fig.\ref{fig:real-world experiment 1}, with the generalized pose trajectory, Panda can successfully pour water into the three cups on the table with a similar trajectory shape.
\label{subsec:pouring water}

\begin{figure}[hbt!]
    \centering
    \includegraphics[trim=0.5cm 1.0cm 0.5cm 0.5cm,clip,width=1\linewidth]{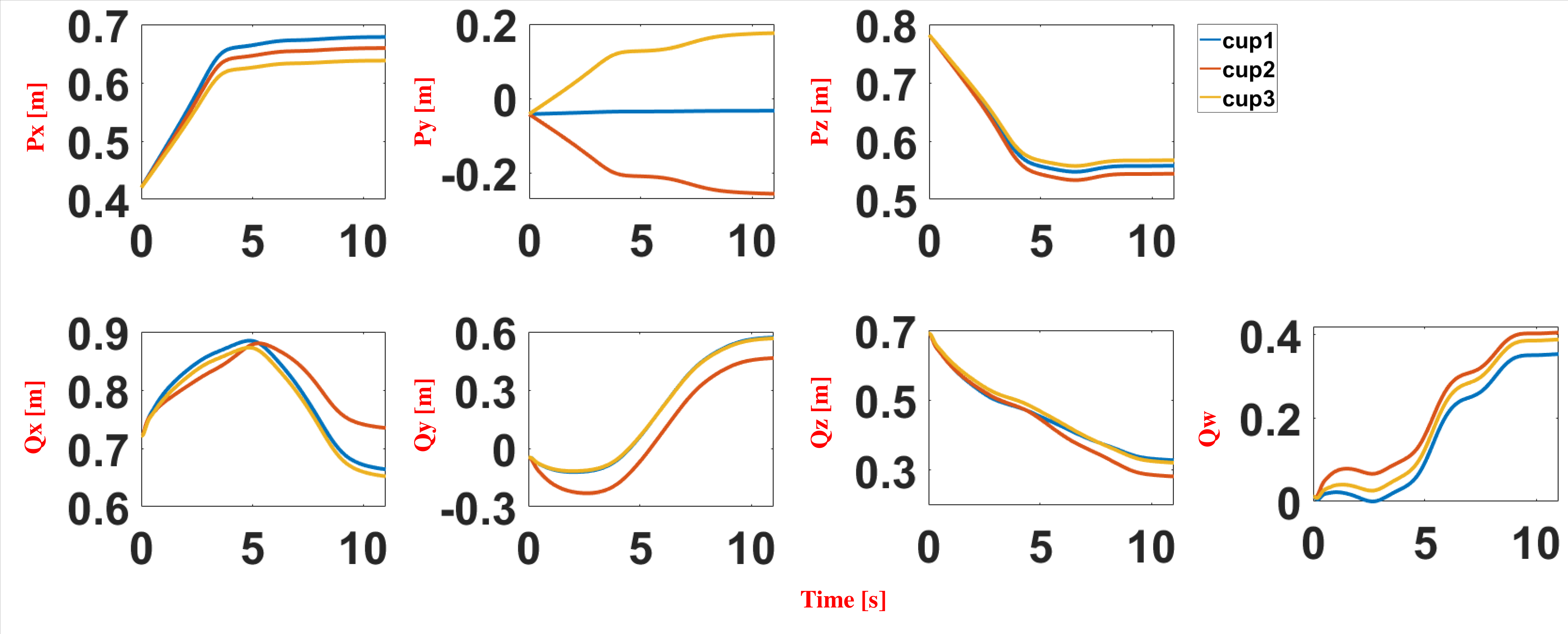}
    \caption{Generalized pose trajectories for pouring wine task}
    \label{fig:pourwine}
    \vspace{-0.3cm}
\end{figure}

Then, we replaced the light plastic bottle with a heavier glass wine bottle. As the wine bottle is longer than the plastic one, this will require different goal poses for pouring wine into the same cups. The generalized pose trajectories for pouring wine are shown in Fig.\ref{fig:pourwine}. 

\begin{figure}[htb!]
\centering
 \includegraphics[trim=0.3cm 0.5cm 0.5cm 0.5cm,clip,width=1\linewidth]{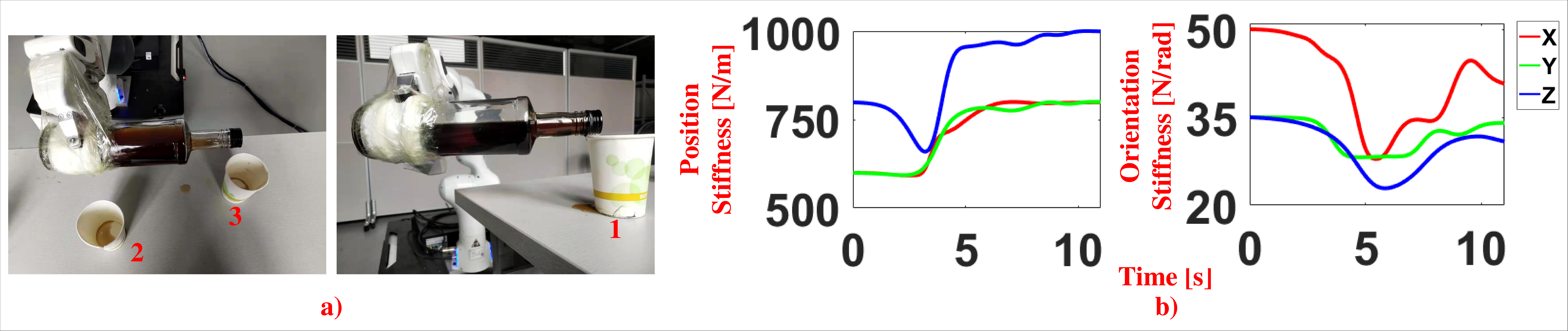}
    \caption{a) Performance of the reference pouring water stiffness profiles in pouring wine task. Panda did not reach the range for pouring wine and even crushed the cup. b) Generalized stiffness profiles that accomplished the pouring wine task}
    \label{fig:real-world_experiment2}
    \vspace{-0.3cm}
\end{figure}

When we executed the new pose trajectories with the reference stiffness profiles for pouring water, Panda failed to pour wine into the first cup and the third cup, and managed to pour wine into the second cup only once, shown in Fig.\ref{fig:real-world_experiment2} a). The main reasons for this failure are that: 1) the glass bottle is heavier than the plastic one. The reference translational stiffness profiles should be increased to compensate for the mass change of the robot end-effector, particularly the stiffness in Axe Z; 2) A longer bottle will enlarge the positional distance errors between the cup and the bottleneck when there are orientational errors. Besides, a heavier bottle also introduced larger external torques at the end-effector. Therefore, the values of rotational stiffness should be increased correctly to compensate for the external torques and to reduce the orientational tracking errors. Meanwhile, no matter it is pouring water or pouring wine, the overall stiffness shape should be kept, as the task constraints did not change for the pouring tasks.

Therefore, we generalized both the stiffness profiles to new goal values and keep its overall tendency with Eq.\eqref{con:dmp_stiffness}. We re-run the pouring wine tests for 3 times. Panda successfully adapted the changes and managed to pour wine into the cups, just like what we have shown in Fig.\ref{fig:real-world experiment 1}

\section{Discussion}
\label{sec: Discussion }
It should be emphasized that our stiffness estimation method estimates both translational and rotational stiffness profiles, while most previously presented methods can not achieve this goal. Another advantage of our method is its efficiency and effectiveness while estimating stiffness profiles. In this paper, with only 8 collected trajectories, we could generate reasonable stiffness profiles to reproduce the demonstrated skill and to further generalize it to new scenarios.

Besides, in this article, we mainly focus on learning and generalizing stiffness profiles from human demonstrations. However, the estimated stiffness profiles may not be optimal for the target task. Improving the stiffness profiles with optimization methods, like PI$^2$\cite{buchli2011learning}, can be useful to further improve the performance of the learned variable impedance manipulation skill in solving the target task.

\section{Conclusion}
\label{sec: Conclusion}
In this work, we proposed an efficient DMP-based imitation learning framework for learning and generating variable impedance manipulation skills from human demonstrations in Cartesian space. This framework not only estimates both translational and rotational stiffness profiles from demonstrated trajectories, but also improves robots' adaptability to environment changes(i.e.the weight and shape changes of the robot's end-effector) by generalizing generated stiffness profiles. The experimental study validates the effectiveness our proposed framework. Besides, we believe it can be used on robots with different configurations, as our framework learns and generalizes skills in Cartesian space.

For future work, we will test our proposed approach in more complex human-robot interaction tasks. It is also an interesting direction to further optimize the generated reference trajectory and stiffness profiles through reinforcement learning algorithms.

\bibliography{main}
\bibliographystyle{unsrt}

\end{document}